\pdfoutput=1

\documentclass[11pt]{article}

\usepackage[final]{acl}

\usepackage{times}
\usepackage{latexsym}

\usepackage[T1]{fontenc}

\usepackage[utf8]{inputenc}

\usepackage{microtype}

\usepackage{inconsolata}

\usepackage{graphicx}
\usepackage{mathtools}
\usepackage{amsmath}
\usepackage{amssymb}
\usepackage{amsfonts}
\usepackage{multirow}
\usepackage[ruled,noline]{algorithm2e}
\usepackage{subcaption}
\usepackage{graphicx}
\usepackage{verbatim}
\usepackage{cite} 

\title{A Generic Method for Fine-grained Category Discovery in Natural Language Texts}
\author{Chang Tian$^{\dagger}$, Matthew B. Blaschko$^{\dagger}$, Wenpeng Yin, Mingzhe Xing, \\  \textbf{Yinliang Yue}, \textbf{Marie-Francine Moens}$^{\dagger}$\\
  $^{\dagger}$ KU Leuven\\
\texttt{chang.tian@kuleuven.be}}
\begin{document}
\maketitle
\begin{abstract}
Fine-grained category discovery using only coarse-grained supervision is a cost-effective yet challenging task. Previous training methods focus on aligning query samples with positive samples and distancing them from negatives. They often neglect intra-category and inter-category semantic similarities of fine-grained categories when navigating sample distributions in the embedding space.
Furthermore, some evaluation techniques that rely on pre-collected test samples are inadequate for real-time applications. To address these shortcomings, we introduce a method that successfully detects fine-grained clusters of semantically similar texts guided by a novel objective function. The method uses semantic similarities in a logarithmic space to guide sample distributions in the Euclidean space and to form distinct clusters that represent fine-grained categories. We also propose a centroid inference mechanism to support real-time applications. The efficacy of the method is both theoretically justified and empirically confirmed on three benchmark tasks. The proposed objective function is integrated in multiple contrastive learning based neural models. 
Its results surpass existing state-of-the-art approaches in terms of Accuracy, Adjusted Rand Index and Normalized Mutual Information of the detected fine-grained categories. Code and data are publicly available at https://github.com/changtianluckyforever/F-grained-STAR.
\end{abstract}

\section{Introduction}
Fine-grained analysis has drawn much attention in many artificial intelligence fields, e.g., Computer Vision~\citep{chen2018knowledge, wang2024content, park2024fine, li2021paint4poem} and Natural Language Processing~\citep{tian2022anti,ma2023coarse,tian2024fighting,an2024down}, because it can provide more detailed features than coarse-grained data.
\begin{figure}[t]
\includegraphics[width=\columnwidth]{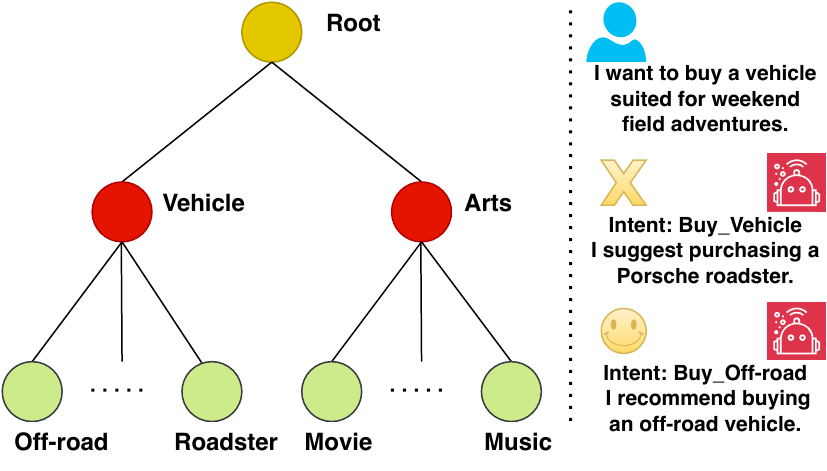}
\centering
\caption{A fine-grained intent detection example. \textbf{Left}: This panel illustrates the label hierarchy, transitioning from coarse-grained to fine-grained granularity. \textbf{Right}: This example demonstrates intent detection in a conversation about car choices, showing how coarse-grained analysis alone can lead to incorrect recommendations by a life assistant due to a lack of fine-grained analysis.}
\label{img:fine_intent}
\end{figure}
For instance, as illustrated in Figure~\ref{img:fine_intent}, solely based on coarse-grained analysis, the chatbot might incorrectly recommend a roadster, which is unsuitable for field adventures. Detecting the fine-grained intent would allow the chatbot to recommend an off-road vehicle that aligns with the user's requirements. However, annotating fine-grained categories can be labor-intensive, as it demands precise expert knowledge specific to each domain and involved dataset.
Addressing this challenge, ~\citet{an2022fine} recently introduced Fine-grained Category Discovery under Coarse-grained Supervision (\textbf{FCDC}) for language classification tasks (details in Section~\ref{sec:problem_formulation}). FCDC aims to reduce annotation costs by leveraging the relative ease of obtaining coarse-grained annotations, without requiring fine-grained supervisory information. This approach has sparked significant research interest in the automatic discovery of fine-grained language categories.~\citep{ma2023coarse, an2023dna, vaze2024no, lian2024learning}.

Existing methods for addressing FCDC are typically grouped into three groups~\citep{an2024down}: language models, self-training methods, and contrastive learning methods. Language models~\citep{devlin2019bert,touvron2023llama}, including their fine-tuned versions with coarse labels, generally perform poorly on this task due to a lack of fine-grained supervision. Self-training methods~\citep{caron2018deep,zhang2021discovering} and their variants often employ clustering assignments as fine-grained pseudo labels, filtering out some noisy pseudo labels, and training with these labels. Dominant contrastive learning methods~\citep{chen2020simple,mekala2021coarse2fine,an2022fine,an2023dna} typically identify positive and negative samples for a given query by measuring their semantic distances. The contrastive loss ensures that the query sample moves closer to positive samples and further away from negative samples. So these methods form clusters of samples in the embedding space, with each cluster representing a discovered fine-grained category, without requiring fine-grained category supervision.

\begin{figure}[t] 
\includegraphics[width=0.5\columnwidth]{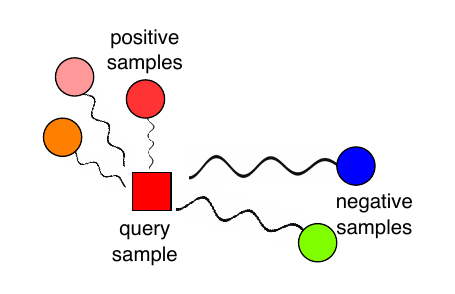}
\centering
\caption{Visualization of comprehensive semantic similarities (CSS). The wavy line indicates the bidirectional KL divergence between two samples. }
\label{img:comprehensive_similarity}
\end{figure}

However, past methods did not utilize \textbf{comprehensive semantic similarities} (\textbf{CSS}) in the logarithmic space to guide sample distributions in Euclidean space. We define CSS as the fine-grained semantic similarities measured by bidirectional KL divergence in the logarithmic space between the query sample and each available positive or negative sample, as illustrated in Figure~\ref{img:comprehensive_similarity}. Although~\citet{an2024down} recently explored similarities measured by rank order between the query sample and positive samples, they ignore similarities with negative samples.

We propose a method (\textbf{STAR}) for detecting fine-grained clu\textbf{st}ers of semantically simil\textbf{ar} texts through a novel objective function, with the core component considering CSS. This component guides sample distributions in the Euclidean space based on the magnitude of CSS in the logarithmic space. Large semantic differences (low similarity) in the logarithmic space between the query sample and an available sample push the query sample further away in Euclidean space, while small semantic differences bring the query sample closer to the available sample. Thus, samples form distinguishable fine-grained clusters in Euclidean space, with each cluster representing a discovered category.

Additionally, clustering inference used by previous works~\citep{an2022fine, an2023dna, an2024down} can not support real-time scenarios, so we propose a variant inference mechanism utilizing approximated fine-grained cluster centroids, delivering competitive results for the tasks considered.

Our main contributions in this work can be summarized as follows:
\begin{itemize}
    \item \textbf{Method:} 
    STAR enhances existing contrastive learning methods by leveraging comprehensive semantic similarities in a logarithmic space to guide sample distributions in the Euclidean space, thereby making fine-grained categories more distinguishable.
    \item \textbf{Theory:} We interpret STAR from the perspectives of clustering and generalized Expectation Maximization (EM). Also, we conduct loss and gradient analyses to explain the effectiveness of using CSS 
    for category discovery.
    \item \textbf{Experiments:} Experiments on three text classification tasks (intent detection~\citep{larson2019evaluation}, scientific abstract classification~\citep{kowsari2017hdltex}, and chatbot query~\citep{liu2021benchmarking}) demonstrate new state-of-the-art (SOTA)  performance compared to 22 baselines, validating the theoretical method.
\end{itemize}
\section{Related Work}
\subsection{Fine-grained Category Discovery}
Fine-grained data analysis is crucial in Natural Language Processing~\citep{guo2021overview, ma2023coarse, tian2024fighting} and Computer Vision~\citep{pan2023fine, wang2024accurate}. However, effectively discovering fine-grained categories from coarse-grained ones remains  challenging~\citep{mekala2021coarse2fine}. Traditional category discovery methods often assume that known and discovered categories are at the same granularity level~\citep{an2023generalized, vaze2024no}.

To discover fine-grained categories under the supervision of coarse-grained categories, \citep{an2022fine} introduced the FCDC task. Self-training approaches, such as Deep Cluster~\citep{caron2018deep, an2023dna}, use clustering algorithms to detect the fine-grained categories, assign pseudo labels to the clusters and their samples, and then train a classification model with these pseudo labels. Its variant, Deep Aligned Clustering~\citep{zhang2021discovering}, devises a strategy to filter out inconsistent pseudo-labels during clustering. Contrastive learning has become prevalent in FCDC tasks; \citep{bukchin2021fine,an2022fine} developed angular contrastive learning tailored for fine-grained classification. \citet{an2022fine} proposed a weighted self-contrastive framework to enhance the model's discriminative capacity for coarse-grained samples. \citep{ma2023coarse} and \citep{an2023dna} used noisy fine-grained centroids and retrieved neighbors as positive pairs, respectively, applying constraints to filter noise. \citet{an2024down} advanced this approach with 
neighbors that are manually weighted as positive pairs. However, previous efforts have not leveraged comprehensive semantic similarities to guide sample distributions and thereby to 
enhance fine-grained category discovery.
\subsection{Neighborhood Contrastive Learning}
Contrastive learning enhances representation learning by bringing the query sample closer to positive samples and distancing it from negative samples~\citep{chen2020simple}. Prior research has focused on constructing high-quality positive pairs. \citep{he2020momentum} utilized two different transformations of the same input as query and positive sample, respectively. \citet{li2020prototypical} introduced the use of prototypes, derived through clustering, as positive instances. Additionally, \citet{an2022fine} employed shallow-layer features from BERT as positive samples and introduced a weighted contrastive loss. This approach primarily differentiates data at a coarse-grained level, and the manually set weights limit its broader applicability.

To circumvent complex data augmentation, neighborhood contrastive learning (NCL) was developed, treating the nearest neighbors of queries as positive samples~\citep{dwibedi2021little}. \citet{zhong2021neighborhood} extended this by utilizing k-nearest neighbors to identify hard negative samples, while \citet{zhang2022new} selected a positive key from the k-nearest neighbors for contrastive representation learning. However, these approaches often deal with noisy nearest neighbors that include false-positive samples. \citep{an2023dna} addressed this by proposing three constraints to filter out uncertain neighbors, yet they overlooked semantic similarities between query sample and each available sample. \citet{an2024down} represented semantic similarities using rank order among positive samples but neglected similarities among negative samples. In contrast, STAR uses comprehensive semantic similarities to guide sample distributions in the Euclidean space, offering richer features and a superior approach to pure contrastive learning.

\section{Problem Formulation}
\label{sec:problem_formulation}
Given a set of coarse-grained categories $Y_{coarse} = \{C_1, C_2, \dots, C_M\}$ and a coarsely labeled training set $D_{train} = \{(x_i, c_i) \mid c_i \in Y_{coarse}\}_{i=1}^N$, where $N$ denotes the number of training samples, the task of FCDC involves developing a feature encoder $F_{\theta}$. This encoder maps samples into a feature space, further segmenting them into distinct fine-grained categories $Y_{fine} = \{F_1, F_2, \dots, F_K\}$, without any fine-grained supervisory information. Here, $Y_{fine}$ represents sub-classes of $Y_{coarse}$. Model effectiveness is evaluated on a testing set $D_{test} = \{(x_i, y_i) \mid y_i \in Y_{fine}\}_{i=1}^L$, with $L$ as the number of test samples, utilizing features extracted by $F_{\theta}$. For evaluation consistency and fairness, only the number of fine-grained categories $K$ is used, aligning with methodologies established in previous research \citep{ma2023coarse,an2022fine,an2023dna}.
\section{Method}
\begin{figure*}[t!]\includegraphics[width=0.92\textwidth]{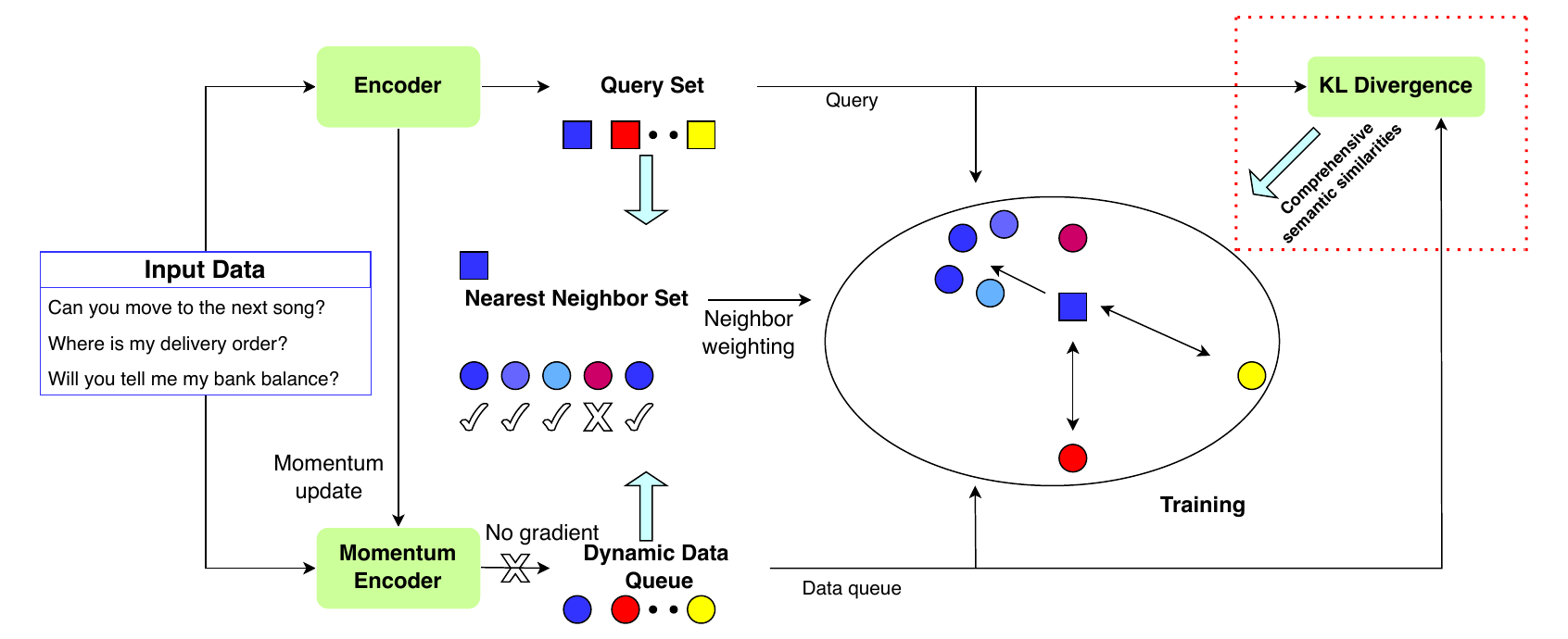}
\centering
\caption{STAR-DOWN integrates the baseline DOWN with the STAR method (shown in the red dashed box). In the visual representation, colors differentiate samples, squares represent features extracted by the Encoder, and circles denote features extracted by the Momentum Encoder. Unidirectional arrows indicate proximity, while bidirectional arrows signify distance between samples.}
\label{img:star_framework}
\end{figure*}
STAR leverages comprehensive semantic similarities and integrates seamlessly with contrastive learning baselines by modifying the objective function. We have developed variants for three baselines: PseudoPrototypicalNet (PPNet)~\citep{boney2017semi,ji2020unsupervised}, DNA~\citep{an2023dna}, and DOWN~\citep{an2024down}. 
This section focuses on STAR-DOWN because DOWN outperforms other baselines, with additional method variants detailed in Appendix

DOWN involves three steps: pre-training with coarse-grained labels (Section~\ref{subsection:step1}), retrieving and weighting nearest neighbors (Section~\ref{subsection:step2}), and training with a contrastive loss. STAR-DOWN follows the same first two steps but replaces the third with a novel objective function (Section~\ref{subsection:step3}).
Like DOWN, STAR-DOWN iterates the last two steps until the unsupervised metric, the silhouette score of the clustering into fine-grained clusters, does not improve for five consecutive epochs. The detailed algorithm is provided in Appendix
\subsection{Multi-task Pre-training}
\label{subsection:step1}
As illustrated in Figure~\ref{img:star_framework}, the baseline DOWN~\citep{an2024down} utilizes the BERT Encoder \( F_{\theta} \) to extract normalized feature embeddings \( q_i = F_{\theta}(x_i) \) for input \( x_i \), where \( \theta \) represents the Encoder parameters. To ensure effective initialization for fine-grained training, DOWN pre-trains the Encoder on the coarsely labeled train set \( D_{train} \) with labels \( Y_{coarse} \). DOWN utilizes the sum of a cross-entropy loss $L_{\text{ce}}$ and a masked language modeling loss $L_{\text{mlm}}$ for multi-task pre-training of the Encoder (detailed in Appendix

\subsection{Neighbors Retrieval and Weighting}
\label{subsection:step2}
In Figure~\ref{img:star_framework}, the Momentum Encoder \( F_{\theta_k} \) with parameters \( \theta_k \) extracts and stores gradient-free normalized neighbor features \( h_i = F_{\theta_k}(x_i) \) in a dynamic data queue \( Q \). To ensure consistency between the outputs of \( F_{\theta_k} \) and \( F_{\theta} \), \( F_{\theta_k} \)'s parameters are updated via a moving-average method~\citep{he2020momentum}:
\(\theta_k \leftarrow m \theta_k + (1 - m) \theta\),
where \( m \) is the momentum coefficient. For each query feature \( q_i \), in order to facilitate semantic similarity capture and fine-grained clustering, its top-k nearest neighbors $N_i$ are determined from \( Q \) using cosine similarity (Sim):
\( N_i = \{h_j \mid h_j \in \operatorname{argtopK}_{h_l \in Q}  (\operatorname{Sim}(q_i, h_l))\} \),
where $\text{Sim}(q_i, h_l) = \frac{q_i^\mathrm{T} h_l}{\|q_i\| \cdot \|h_l\|}$ is the cosine similarity function.

To counteract potential false positives in \( N_i \), DOWN utilizes a soft weighting mechanism based on neighbor rank to balance information utility against noise, with weights \( \omega_j \) of neighbor $h_j$ calculated as:
$\omega_j = \phi \cdot \alpha^{-\frac{l_{ij}}{k}}$,
where \( \phi \) is a normalizing constant for weights, \( \alpha \) serves as the exponential base, \( k \) is the retrieved neighbor count, and \( l_{ij} \) denotes the rank of \( h_j \) as a neighbor to \( q_i \).

To align with the model’s evolving accuracy in neighbor retrieval during training, DOWN periodically decreases \( \alpha \) every five epochs, the values for \( \alpha \) in \( \omega_j \) are: \(\alpha_{\text{set}} = \{150, 10, 5, 2\}\). The \( \omega_j \) of each positive sample \( h_j \) is used in Eqs. \ref{eq:l1_loss} and \ref{eq:l2_loss}.

\subsection{Training}
\label{subsection:step3}
\subsubsection{Objective Function}
Given a training batch \( N_{train} \in D_{train} \), where \( Y_c \) is the set of coarse-grained labels of \( N_{train} \), DOWN trains the model using the loss:

\begin{equation}
L_{\text{train}} = L_{\text{ce}} + L_{\text{DOWN}},
\end{equation}

\begin{equation}
L_{\text{DOWN}} = \frac{1}{|N_{train}|} \sum_{q_i \in N_{train}} L_{1}^i.
\end{equation}

As shown in Eq.~\ref{eq:l1_loss}, DOWN uses a conventional contrastive objective function in the Euclidean space, while STAR-DOWN introduces a novel objective function in Eq.~\ref{eq:l2_loss}, leveraging CSS in a logarithmic space to guide sample distributions in the Euclidean space, the temperature $\tau$ is a fixed constant in Eq.~\ref{eq:l1_loss} and Eq.~\ref{eq:l2_loss}:
\begin{equation}
\begin{aligned}
L_{1}^i = - \sum_{h_j \in N_i} \omega_j \cdot \log \frac{\exp(q_i^\mathrm{T} h_j / \tau)}{\sum\limits_{h_k \in Q} \exp(q_i^\mathrm{T} h_k / \tau)}.
\end{aligned}
\label{eq:l1_loss}
\end{equation}
{\small
\begin{equation}
\begin{aligned}
L_{2}^i = &- \gamma \sum_{h_j \in N_i} \omega_j \cdot \log \frac{\exp(-d_{KL}(q_i, h_j) / \tau)}{\sum\limits_{h_k \in Q} \exp(-d_{KL}(q_i, h_k) / \tau)} \\
&- \sum_{h_j \in N_i} \omega_j \cdot \log \frac{\exp(q_i^\mathrm{T} h_j / \tau)}{\sum\limits_{h_k \in Q} B^{d_{KL}(q_i, h_k)} \cdot \exp(q_i^\mathrm{T} h_k / \tau)}.
\end{aligned}
\label{eq:l2_loss}
\end{equation}
}

During training, STAR-DOWN optimizes the following objective function:
\begin{equation}
L_{\text{train}} = L_{\text{ce}} + L_{\text{STAR}},
\label{eq:star_down_train}
\end{equation}

\begin{equation}
L_{\text{STAR}} = \frac{1}{|N_{train}|} \sum_{q_i \in N_{train}} L_{2}^i.
\end{equation}

As shown in Eq.~\ref{eq:l2_loss}, the term $d_{KL}(q_i, h_k)$ in $L_{2}^i$ represents the bidirectional KL divergence in a logarithmic space between the query sample embedding \( q_i \) and the data queue sample embedding $h_k$ (detailed in Appendix
 $B$ is a trainable scalar representing the exponential base.

The first term in $L_{2}^i$ minimizes the KL divergence between query samples and positive samples (retrieved neighbors in Section~\ref{subsection:step2}) while increasing it for negative samples in the logarithmic space, with $\gamma$ as a balancing hyperparameter. The second term in $L_{2}^i$ uses CSS in the logarithmic space, denoted by $B^{d_{KL}(q_i, h_k)}$, to guide query sample distribution in the Euclidean space. $q_i^\mathrm{T} h_k$ quantifies the cosine similarity between normalized \( q_i \) and \( h_k \), equivalent to the negative Euclidean distance (detailed in Appendix
. The value of the trainable scalar $B$ is updated during loss backpropagation, so $B^{d_{KL}(q_i, h_k)}$ is fully trainable and can integrate with contrastive learning methods, making the STAR method \emph{generic}.
\subsubsection{Loss Analysis}
Since STAR-DOWN discovers fine-grained categories in the Euclidean space, we analyze the second term \( L_{2-2}^i \) of the loss \( L_{2}^i \), which optimizes sample distributions in the Euclidean space:

{\small
\begin{equation}
\begin{aligned}
 L_{2-2}^i =& - \sum_{h_j \in N_i} \omega_j \cdot \log \frac{\exp(q_i^\mathrm{T} h_j / \tau)}{\sum\limits_{h_k \in Q} B^{d_{KL}(q_i, h_k)} \cdot \exp(q_i^\mathrm{T} h_k / \tau)} \\
=& \sum_{h_j \in N_i} \omega_j \cdot ( \log \sum\limits_{h_k \in Q} B^{d_{KL}(q_i, h_k)} \cdot \exp(q_i^\mathrm{T} h_k / \tau) \\
&- (q_i^\mathrm{T} h_j / \tau) ).
\end{aligned}
\label{eq:l2_loss_2}
\end{equation}
}

In the loss \( L_{2-2}^i \), \( B^{d_{KL}(q_i, h_k)} \) uses CSS in the logarithmic space to guide sample distributions in the Euclidean space. A large \( d_{KL}(q_i, h_k) \) (low semantic similarity) causes \( q_i \) to distance itself from \( h_k \) in the Euclidean space, reducing \( q_i^\mathrm{T} h_k \), while a small \( d_{KL}(q_i, h_k) \) allows \( q_i \) to remain relatively close to \( h_k \) compared to negative samples. This results in the formation of compact fine-grained clusters, with each cluster representing a discovered category. We also analyze the STAR method from the perspectives of \textbf{gradient}, \textbf{clustering}, and \textbf{generalized EM}. Detailed analyses are provided in Appendix 

\subsection{Inference}
\label{sec:evaluation_mechanism}
Previous methods~\citep{an2023dna, an2024down} use clustering inference on sample embeddings from \( F_{\theta} \) extracted from \( D_{test} \), which is unsuitable for real-time tasks, such as intent detection, which require immediate response and can not wait to collect enough test samples for clustering. We introduce an alternative, centroid inference, suitable for both real-time and other contexts. Using \( F_{\theta} \), we derive sample embeddings from \( D_{train} \) and assign fine-grained pseudo labels through clustering. For each fine-grained cluster, only the embeddings of samples from the predominant coarse-grained category (the category with the most samples in this fine-grained cluster) are averaged to form centroid representations. These approximated centroids are used to determine the fine-grained category of each test sample based on cosine similarity. A visual explanation is in Appendix
\section{Experiments}
\subsection{Experimental Settings}
\subsubsection{Datasets}
\begin{table}[t!]
\centering
\begin{tabular}{ccccc}
\hline
Dataset & $\lvert \mathcal{C} \rvert$ & $\lvert \mathcal{F} \rvert$ & \# Train & \# Test \\ \hline
CLINC  & 10  & 150 & 18000 & 1000 \\
WOS    & 7   & 33  & 8362  & 2420 \\
HWU64  & 18  & 64  & 8954  & 1031 \\ \hline
\end{tabular}
\caption{Statistics of datasets~\citep{an2023dna}. \#: number of samples. $\lvert \mathcal{C} \rvert$: number of coarse-grained categories. $\lvert \mathcal{F} \rvert$: number of fine-grained categories.}
\label{tab:data_statistics}
\end{table}
We conduct experiments on three benchmark datasets: CLINC~\citep{larson2019evaluation}, WOS~\citep{kowsari2017hdltex}, and HWU64~\citep{liu2021benchmarking}. CLINC is an intent detection dataset spanning multiple domains. WOS is used for paper abstract classification, and HWU64 is designed for assistant query classification. Dataset statistics are provided in Table~\ref{tab:data_statistics}.
\subsubsection{Baselines for Comparison}
We compare our methods against the following baselines. \textbf{Language models}: BERT~\citep{devlin-etal-2019-bert}, BERT with coarse-grained fine-tuning, Llama2~\citep{touvron2023llama}, Llama2 with coarse-grained fine-tuning and  GPT4~\citep{achiam2023gpt}. \textbf{Self-training} baselines: DeepCluster (DC)~\citep{caron2018deep}, DeepAlignedCluster (DAC)~\citep{zhang2021discovering}, and PseudoPrototypicalNet (PPNet)~\citep{boney2017semi,ji2020unsupervised}. \textbf{Contrastive learning} baselines: SimCSE~\citep{gao2021simcse}, Ancor~\citep{bukchin2021fine}, Delete~\citep{wu2020clear}, Nearest-Neighbor Contrastive Learning (NNCL)~\citep{dwibedi2021little}, Contrastive Learning with Nearest Neighbors (CLNN)~\citep{zhang2022new}, Soft Neighbor Contrastive Learning (SNCL)~\citep{chongjian2022soft}, Weighted Self-Contrastive Learning (WSCL)~\citep{an2022fine}, Denoised Neighborhood Aggregation (DNA), and Dynamic Order Weighted Network (DOWN)~\citep{an2023dna,an2024down}. We also explore variants incorporating the cross-entropy loss (+CE). 
\subsubsection{Evaluation Metrics}
To evaluate the quality of the discovered fine-grained clusters, we use the Adjusted Rand Index (ARI)~\citep{hubert1985comparing} and Normalized Mutual Information (NMI)~\citep{lancichinetti2009detecting}. For assessing classification performance, we use clustering Accuracy (ACC)~\citep{kuhn1955hungarian,an2023dna}. Detailed descriptions of these metrics are provided in Appendix
\subsubsection{Implementation Details}
To ensure fair comparisons with baselines, we use the BERT-base-uncased model as the backbone for all STAR method variants. We adhere to the hyperparameters used by the integrated baselines to demonstrate the effectiveness of our STAR method. The learning rate for both pre-training and training is \(5e^{-5}\), using the AdamW optimizer with a 0.01 weight decay and 1.0 gradient clipping. The momentum coefficient $m$ is set to 0.99. The batch size for pre-training, training, and testing is 64. The temperature \(\tau\) is set to 0.07. The number of neighbors \(k\) is set to \{120, 120, 250\} for the CLINC, HWU64, and WOS datasets, respectively. Epochs for pretraining and training are set to 100 and 20, respectively. Further details are provided in Appendix
\subsubsection{Research Questions}
The following research questions (\textbf{RQ}s) are investigated: 
1. What is the impact of STAR method on FCDC tasks?
2. What are the effects of the proposed real-time centroid inference compared to traditional clustering inference?
3. How does each component of the STAR method affect performance?
4. How can we effectively and efficiently set the base for the exponential function in the STAR method?
\subsection{Result Analysis (RQ1)}
\begin{table*}[t]
\centering
\resizebox{\textwidth}{!}{
\begin{tabular}{lccc|ccc|ccc}
\hline
\multirow{2}{*}{\textbf{Methods}} & \multicolumn{3}{c|}{\textbf{HWU64}} & \multicolumn{3}{c|}{\textbf{CLINC}} & \multicolumn{3}{c}{\textbf{WOS}} \\ \cline{2-10} 
                         & ACC   & ARI   & NMI   & ACC   & ARI   & NMI   & ACC   & ARI   & NMI   \\ \hline
BERT~\citep{devlin-etal-2019-bert}           & 33.52 & 17.04 & 56.90 & 34.37 & 17.61 & 64.75 & 31.97 & 18.36 & 45.15 \\
BERT + CE      & 37.89 & 33.68 & 74.63 & 43.85 & 32.37 & 78.58 & 38.29 & 36.94 & 64.72 \\ 
Llama2~\citep{touvron2023llama}      & 19.27±1.21 & 5.21±0.46 & 44.34±0.85 & 20.77±2.61 & 5.83±1.52 & 49.7±3.68 & 9.85±1.14 & 1.26±0.75 & 18.27±2.28 \\ 
Llama2 + CE      & 32.40±5.46 & 17.32±5.95 & 57.53±5.78 & 45.69±6.85 & 29.38±6.55 & 72.66±7.13 & 18.51±1.50 & 7.8±1.18 & 29.66±3.23 \\ 
GPT4~\citep{achiam2023gpt}      & 10.77±1.86 & 0.14±0.05 &  35.17±3.68 & 9.56±2.12 & 0.11±0.06 & 46.69±3.24 & 7.56±1.51 & 0.15±0.04  & 27.78±2.98 \\ \hline
DC~\citep{caron2018deep}               & 18.05 & 43.34 & 29.74 & 26.40 & 12.51 & 61.26 & 29.17 & 13.98 & 53.27 \\
DAC~\citep{zhang2021discovering}            & 29.14 & 12.89 & 52.99 & 29.16 & 14.15 & 62.78 & 28.47 & 15.94 & 43.52 \\
DC + CE          & 41.73 & 27.81 & 66.81 & 30.28 & 13.56 & 62.38 & 38.76 & 35.21 & 60.30 \\
DAC + CE        & 42.19 & 28.15 & 66.50 & 42.09 & 28.09 & 72.78 & 39.42 & 33.67 & 61.60 \\ 
PPNet~\citep{ji2020unsupervised}      &58.36±2.51  & 47.63±1.96  & 79.75±1.02   &     70.15±1.86 &   59.31±0.96 &   85.08±0.81   &    62.59±1.41 &  50.81±1.21 &  72.19±0.68   \\
\textbf{STAR-PPNet} (ours)   &63.19±2.38  & 52.21±1.33 &  81.66±1.21     &   73.21±1.97  &  61.87±0.79  &  86.16±0.47   &   66.15±1.33   &53.61±1.24  &73.82±0.74    \\ \hline
Delete~\citep{wu2020clear}       & 21.30 &  6.52 & 44.13 & 47.11 & 31.28 & 73.39 & 24.50 & 11.68 & 35.47 \\
SimCSE~\citep{gao2021simcse}       & 24.48 &  8.42 & 46.94 & 40.22 & 23.57 & 69.02 & 25.87 & 13.03 & 38.53 \\
Ancor + CE    & 32.90 & 30.71 & 74.73 & 44.44 & 31.50 & 74.67 & 39.34 & 26.14 & 54.35 \\
NNCL~\citep{dwibedi2021little}           & 32.98 & 30.02 & 73.24 & 17.42 & 13.93 & 67.56 & 29.64 & 28.51 & 61.37 \\
SimCSE + CE   & 34.04 & 31.81 & 74.86 & 52.53 & 37.03 & 77.39 & 41.28 & 34.47 & 61.62 \\
Delete + CE   & 35.13 & 31.84 & 74.88 & 47.87 & 33.79 & 76.25 & 41.53 & 33.78 & 61.01 \\
CLNN~\citep{zhang2022new}           & 37.21 & 34.66 & 75.27 & 19.96 & 14.76 & 68.30 & 29.48 & 28.42 & 60.99 \\
Ancor~\citep{bukchin2021fine}         & 37.34 & 34.75 & 74.99 & 45.60 & 33.11 & 75.23 & 41.20 & 37.00 & 65.42 \\
SNCL~\citep{chongjian2022soft}           & 42.32 & 38.17 & 76.39 & 55.01 & 45.64 & 82.93 & 36.27 & 33.62 & 62.35 \\
WSCL~\citep{an2022fine}           & 59.52 & 49.34 & 79.31 & 74.02 & 62.98 & 88.37 & 65.27 & 51.78 & 72.46 \\
DNA~\citep{an2023dna}             & 70.81 & 59.66 & 83.31 & 87.66 & 81.82 & 94.69 & 74.57 & 63.30 & 76.86 \\ 
\textbf{STAR-DNA} (ours)           & 75.79±0.93 & 65.27±1.12 & 85.34±0.36 & 89.25±0.17 & 83.47±0.27 & 95.11±0.05 & 77.19±0.81 & 64.97±0.75 & 77.91±0.76 \\ 
DOWN~\citep{an2024down}                      & 78.92 & 68.17 & 86.22 & 91.79 & 86.70 & 96.05 & 80.00 & 67.09 & 78.87 \\ 
\textbf{STAR-DOWN} (ours)                    & \textbf{80.31±0.26} & \textbf{70.22±0.59} & \textbf{87.28±0.31} & \textbf{92.45±0.38} & \textbf{87.05±0.17} & \textbf{96.20±0.07} & \textbf{81.98±0.67} & \textbf{69.27±0.60} & \textbf{79.99±0.40} \\ \hline
\end{tabular}
}
\caption{The average performance (\%) in terms of Accuracy (ACC), Adjusted Rand Index (ARI), and Normalized Mutual Information (NMI) on three datasets for the FCDC language task. To ensure fair comparisons with previous works~\citep{an2022fine, an2023dna, an2024down} and demonstrate the effectiveness of STAR, we use the same clustering inference mechanism and also average the results over three runs with identical common hyperparameters. Some baselines results are cited from aforementioned previous works, where standard deviations are not originally provided.}
\label{tab:averaged_performance}
\end{table*}

\begin{table*}[t]
\centering
\resizebox{\textwidth}{!}{%
\begin{tabular}{lccc|ccc|ccc}
\hline
\textbf{Methods} & \multicolumn{3}{c|}{\textbf{HWU64}} & \multicolumn{3}{c|}{\textbf{CLINC}} & \multicolumn{3}{c}{\textbf{WOS}} \\ \cline{2-10} 
                         & ACC   & ARI   & NMI   & ACC   & ARI   & NMI   & ACC   & ARI   & NMI   \\ \hline
STAR-DOWN (clustering)                & 80.31±0.26 & 70.22±0.59 & 87.28±0.31 & 92.45±0.38 & 87.05±0.17 & 96.20±0.07 & 81.98±0.67 & 69.27±0.60 & 79.99±0.40 \\
STAR-DOWN (centroid)                     & 79.44±0.51 & 69.13±0.75 & 86.97±0.40 & 92.60±0.45 & 87.16±0.53 & 96.21±0.09 & 81.89±0.53 & 69.05±0.39 & 79.78±0.32 \\ \hline
\end{tabular}%
}
\caption{Comparison of clustering and centroid inference mechanisms. "Clustering" clusters test set sample embeddings to determine each sample's fine-grained category, while "Centroid" infers the category by comparing each test sample's cosine similarity to fine-grained centroids.}
\label{tab:evaluation_mechanism}
\end{table*}
As shown in Table~\ref{tab:averaged_performance}, STAR method variants outperform SOTA methods across all datasets and metrics, validating the effectiveness of the STAR method in FCDC tasks. Language models like BERT, Llama2 and GPT4~\citep{devlin-etal-2019-bert,touvron2023llama,achiam2023gpt} (GPT4 prompt in Appendix
) perform poorly on the FCDC task due to the lack of fine-grained supervisory information.
Self-training methods like DC, DAC, and PPNet~\citep{caron2018deep, zhang2021discovering,ji2020unsupervised} also struggle because they rely on noisy fine-grained pseudo-labels and overlook comprehensive semantic similarities (CSS). Contrastive learning methods such as SNCL~\citep{chongjian2022soft} and WSCL~\citep{an2022fine} perform better by leveraging positive pairs. DNA~\citep{an2023dna} and DOWN~\citep{an2024down} further enhance feature quality by filtering false positives and weighting them by rank. However, these methods still do not use CSS for sample distributions. Integrating the STAR method with existing baselines enhances performance across all datasets, consistently improving sample distributions in Euclidean space.

The superior performance of STAR is attributed to three factors: First, bidirectional KL divergence measures CSS, pushing negative samples further away and relatively bringing positive samples closer based on CSS magnitude, making fine-grained clusters easier to distinguish. Second, the base \(B\) of the exponential in Eq.~\ref{eq:l2_loss} is a trainable scalar, balancing CSS magnitude and semantic structure. Third, STAR variants iteratively bootstrap model performance in neighborhood retrieval and representation learning through a generalized EM process (detailed in Appendix
).

\subsection{Inference Mechanism Comparison (RQ2)}
Previous methods~\citep{chongjian2022soft, an2023dna, an2024down} perform a nearest neighbor search over the examples of the found fine-grained clusters 
for fine-grained category prediction (we refer to this technique as cluster inference). 
We speed up this process making it better suitable for real-time tasks by developing a centroid inference mechanism (see Section~\ref{sec:evaluation_mechanism}). Results in Table~\ref{tab:evaluation_mechanism} demonstrates that results of centroid inference are competitive with cluster inference. When results are of the former are lower, this is
due to two factors: clustering inference leverages inter-relations among test set samples for richer features, while centroid inference depends on centroids derived from noisy samples with fine-grained pseudo-labels. Despite these issues, centroid inference remains a viable option for real-time applications, balancing immediate analytical needs with slight performance trade-offs.
\subsection{Ablation Study (RQ1 \& RQ3)}
\begin{table}[t]
\centering
\label{tab:ablation}
\resizebox{\columnwidth}{!}{%
\begin{tabular}{lccc}
\hline
\textbf{Methods} & \textbf{ACC} & \textbf{ARI} & \textbf{NMI} \\ \hline
\textbf{ours} & 80.31±0.26  & 70.22±0.59 & 87.28±0.31 \\
w/o CE & 78.61±0.44  & 67.32±0.86  & 85.62±0.36  \\ 
w/o KL loss  & 78.97±0.32  & 68.03±0.36  & 85.81±0.16 \\ 
w/o KL weight  & 79.26±0.42 & 68.86±0.37 & 86.21± 0.07 \\ 
w/o KL weight and loss  & 78.96±0.15 & 68.21±0.22 & 86.32±0.10  \\ \hline
\end{tabular}%
}
\caption{Results (\%) of the ablation study for STAR-DOWN on the HWU64 Dataset.}
\label{tab:ablation_study}
\end{table}
We examine the impact of various components of the STAR method in STAR-DOWN, as detailed in Table~\ref{tab:ablation_study}. Our results yield the following insights. (\textbf{1}) Excluding coarse-grained supervision information during training (w/o CE) reduces model performance, as this information is crucial for effective representation learning.
(\textbf{2}) Omitting the first loss term (w/o KL loss) from Eq.~\ref{eq:l2_loss} diminishes performance. The KL loss term aligns the KL divergence between data samples and the query with their semantic similarities. Without it, $B^{d_{KL}(q_i, h_k)}$ fails to guide the query sample distribution based on semantic similarities in Eq.~\ref{eq:l2_loss}. (\textbf{3}) Removing the KL weight $B^{d_{KL}(q_i, h_k)}$ from Eq.~\ref{eq:l2_loss} (w/o KL weight) reduces effectiveness. The loss no longer utilizes fine-grained semantic similarities measured by $B^{d_{KL}(q_i, h_k)}$ in the logarithmic space to direct the query sample distribution in comparison to all samples. (\textbf{4}) Eliminating both the KL loss term and the KL weight in Eq.~\ref{eq:l2_loss} leads to a performance decline. This omission prevents the optimization of the query sample towards positive samples in the logarithmic space and fails to leverage fine-grained semantic similarities in logarithmic space to influence the distribution of query samples relative to all samples in the Euclidean space.
\subsection{Exponential Base Impact (RQ4)}
In the STAR method's loss equation (Eq.~\ref{eq:l2_loss}), $B^{d_{KL}(q_i, h_k)}$ modulates the distribution of $q_i$ and $h_k$ in the Euclidean space based on their semantic similarity in the logarithmic space, as quantified by the bidirectional KL divergence. The base $B$ is used to enhance semantic differences, improving the discriminability of fine-grained categories. We experimented with multiple constant values and a trainable configuration for $B$, with multiple STAR-DOWN results presented in Table~\ref{tab:base_B_selection}. The multiple STAR-DOWN methods with various base values consistently outperform the DOWN method (Table~\ref{tab:averaged_performance}), demonstrating the effectiveness and robustness of the STAR method regardless of the base value \( B \).\footnote{For interesting form similarities to physical laws within the STAR method, see Appendix
.} Notably, base values that are either too low (e.g., $e$) or too high (e.g., 66) disrupt the semantic representation by inadequately or excessively emphasizing semantic similarities in the logarithmic space. To set base value conveniently, we set $B$ as a trainable scalar, achieving favorable outcomes as indicated in Table~\ref{tab:base_B_selection}.
\begin{table}[t]
\centering
\resizebox{\columnwidth}{!}{%
\begin{tabular}{lccc}
\hline
\textbf{Base value} & \textbf{ACC} & \textbf{ARI} & \textbf{NMI} \\ \hline
trainable $B$ (ours) & 80.31±0.26 & 70.22±0.59 & 87.28±0.31 \\
$e$ & 79.96±0.12  & 68.89±0.55 & 86.66±0.10  \\ 
10  & 80.22±0.27 & 69.61±0.65 & 87.08±0.30 \\ 
16  & 80.73±0.32  & 70.14±0.58 & 87.25±0.36 \\ 
66  & 80.57±0.38  & 70.20±0.52 & 87.07±0.15  \\ \hline
\end{tabular}%
}
\caption{Averaged results (\%) and their standard deviations over three runs of multiple STAR-DOWN methods with \textbf{five different base values} on the HWU64 dataset. To set base value conveniently, we set $B$ as a trainable scalar.}
\label{tab:base_B_selection}
\end{table}
\subsection{Inference of Category Semantics}
Prior works~\citep{an2023dna, an2024down} only discovered fine-grained categories and assigned them numeric indices without elucidating the categories semantics, thus constraining their broader application. We propose utilizing the commonsense reasoning capabilities of large language models (\textbf{LLM}s) to infer the semantics of these categories. Specifically, we employ a trained encoder, \( F_{\theta} \), to extract embeddings from all train set samples and cluster these embeddings to assign fine-grained pseudo-labels to each train set sample. For each fine-grained category indicated by a specific pseudo-label, we aggregate all predicted samples from the training set and use an LLM to deduce the category semantics. Details on the LLM prompt are provided in Appendix
.
\subsection{Visualization}
\begin{figure}[t]
\includegraphics[width=0.5\columnwidth]{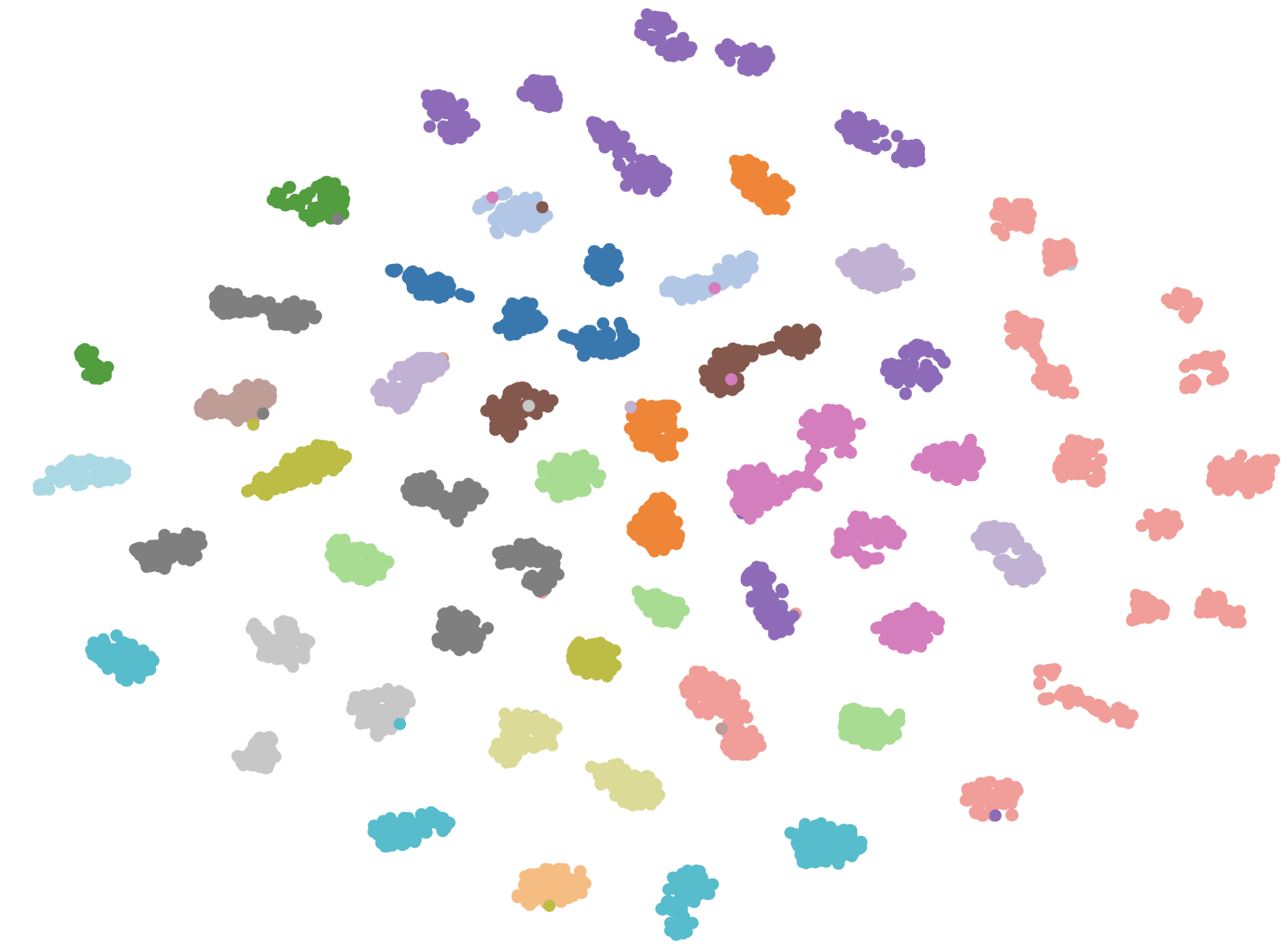}
\centering
\caption{The t-SNE visualization of sample embeddings from STAR-DOWN method on the HWU64 dataset, with different colors representing different coarse-grained categories. The distinct clusters represent the discovered fine-grained categories.}
\label{img:tsne_visualization}
\end{figure}
We visualize the sample embeddings of STAR-DOWN in Figure~\ref{img:tsne_visualization}. The results demonstrate that our method forms distinguishable clusters for fine-grained categories, proving STAR's effectiveness in separating dissimilar samples and clustering similar ones. Additionally, we visualize the generalized EM perspective of STAR-DOWN in Appendix.
\section{Conclusion}
We propose the STAR method for fine-grained category discovery in natural language texts, which utilizes comprehensive semantic similarities in the logarithmic space to guide the distribution of textual samples, including conversational intents, scientific paper abstracts, and assistant queries, in the Euclidean space. STAR pushes query samples further away from negative samples and brings them closer to positive samples based on the comprehensive semantic similarities magnitude. This process forms compact clusters, each representing a discovered category.
We theoretically analyze the effectiveness of STAR method. Additionally, we introduce a centroid inference mechanism that addresses previous gaps in real-time evaluations. Experiments on three natural language benchmarks demonstrate that STAR achieves new state-of-the-art performance in fine-grained category discovery tasks for text classification.

\bibliography{arxivE/acl_latex}

\end{document}